\documentclass[11pt]{article}

\usepackage[final]{acl}

\usepackage{times}
\usepackage{latexsym}

\usepackage[T1]{fontenc}

\usepackage[utf8]{inputenc}

\usepackage{microtype}

\usepackage{inconsolata}

\usepackage{graphicx}

\usepackage{hyperref}
\hypersetup{hidelinks}
\usepackage{amsmath}
\usepackage{booktabs} 
\usepackage{multirow} 
\usepackage{makecell}

\title{EnsemHalDet: Robust VLM Hallucination Detection\\
via Ensemble of Internal State Detectors}

\author{
 \textbf{Ryuhei Miyazato},
 \textbf{Shunsuke Kitada},
 \textbf{Kei Harada}
\\
 The University of Electro-Communications
\\
 \small{
   \textbf{Correspondence:} \href{harada@uec.ac.jp}{harada@uec.ac.jp}
 }
}

\begin{document}
\maketitle

\begin{abstract}
Vision-Language Models (VLMs) excel at multimodal tasks, but they remain vulnerable to hallucinations that are factually incorrect or ungrounded in the input image. Recent work suggests that hallucination detection using internal representations is more efficient and accurate than approaches that rely solely on model outputs. However, existing internal-representation-based methods typically rely on a single representation or detector, limiting their ability to capture diverse hallucination signals. In this paper, we propose EnsemHalDet, an ensemble-based hallucination detection framework that leverages multiple internal representations of VLMs, including attention outputs and hidden states. EnsemHalDet trains independent detectors for each representation and combines them through ensemble learning. Experimental results across multiple VQA datasets and VLMs show that EnsemHalDet consistently outperforms prior methods and single-detector models in terms of AUC. These results demonstrate that ensembling diverse internal signals significantly improves robustness in multimodal hallucination detection\footnote{\url{https://github.com/ryuhei-miyazato/ensemhaldet}}.
\end{abstract}

\section{Introduction}

Vision-Language Models (VLMs) have demonstrated remarkable capabilities in multimodal tasks such as visual question answering (VQA) and image captioning~\cite{NEURIPS2023_6dcf277e, 10.1093/nsr/nwae403, Chen_2024_CVPR, yue-etal-2025-mmmu}. Despite these advancements, they remain vulnerable to hallucinations, where models confidently generate outputs that are either factually incorrect or ungrounded in the input data. This remains a serious challenge from the perspectives of reliability and safety~\cite{chen-etal-2024-unified-hallucination, yang2025mitigating, jiang2025devils}. Therefore, detecting and mitigating hallucinations in VLMs is an important problem.

~\citet{kalai2025language} argue that hallucinations are an inevitable consequence of large language models (LLMs) being optimized for generating plausible responses. While they cannot be fully eliminated, models can abstain from answering when uncertainty is high. Therefore, accurate hallucination detection is crucial for improving overall system reliability, and this work focuses on detection itself.

\begin{figure}[t]
  \centering
  \includegraphics[width=\linewidth]{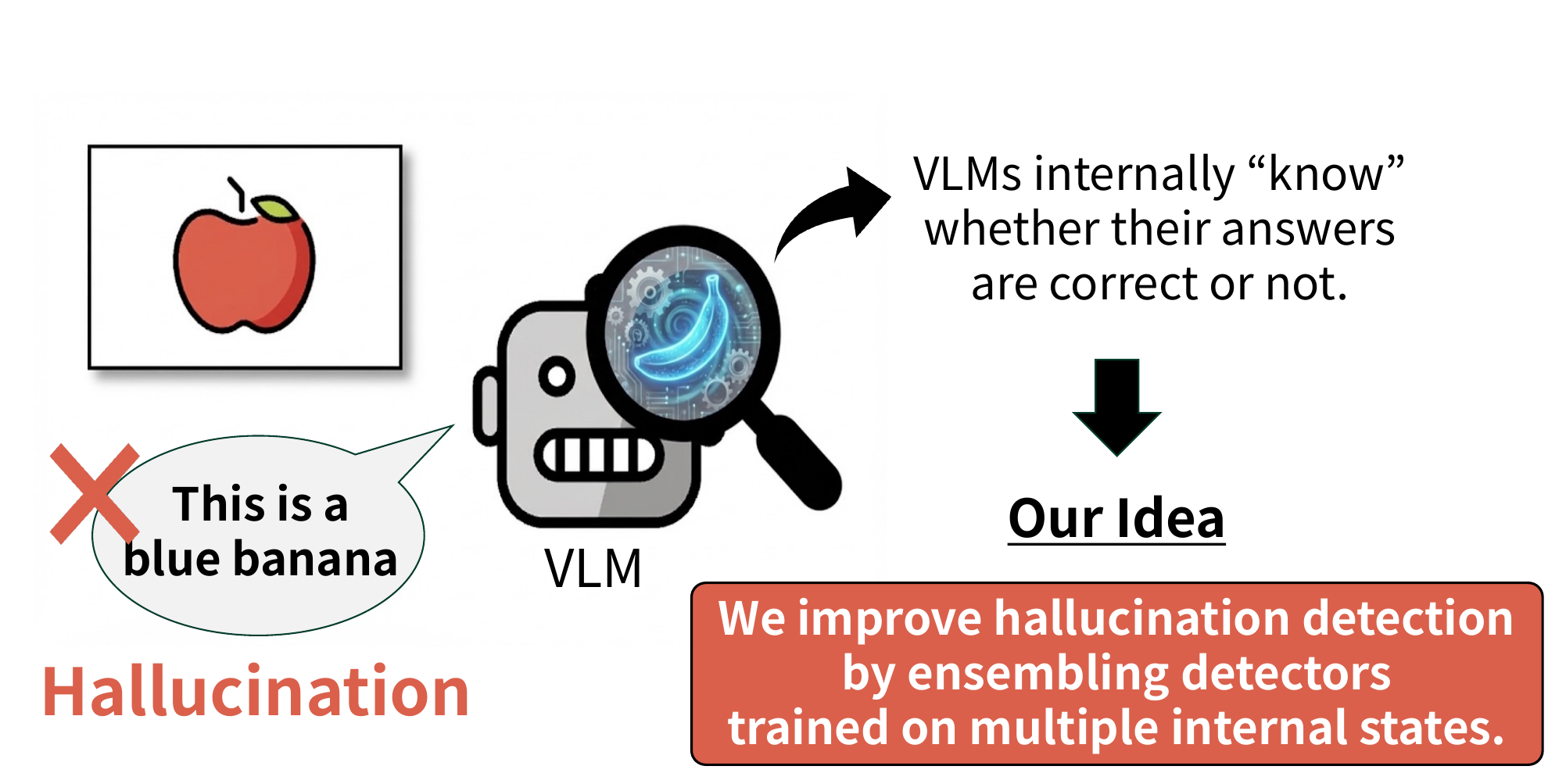}
  \caption{VLMs can produce hallucinated responses that are inconsistent with factual knowledge or image content. However, such hallucinations leave detectable signals in the model’s internal representations. We leverage multiple internal states of VLMs to achieve robust and accurate hallucination detection}
  \label{fig:figure1}
\end{figure}

Prior research on hallucination detection in LLMs / VLMs typically follows three paradigms: (1) output-based methods~\cite{manakul-etal-2023-selfcheckgpt}, which analyze the generated text itself and related indicators, including generation probabilities and self-consistency measures; (2) external-knowledge-based methods~\cite{li-etal-2023-halueval, min-etal-2023-factscore, xue2025verify}, which verify claims against a gold-standard source; and (3) internal-representation-based methods~\cite{azaria-mitchell-2023-internal, su-etal-2024-unsupervised, yang2025mitigating}, which capture signals of hallucination from the model’s internal states. Output-based and external-knowledge-based methods often rely on additional LLM inference or external resources, leading to higher computational costs, yet internal-representation-based approaches achieve better detection performance~\cite{azaria-mitchell-2023-internal}. Internal-representation-based approaches can capture hallucination signals that are not observable in the output, making them a promising direction.

However, most existing methods rely on a single internal representation~\cite{azaria-mitchell-2023-internal, su-etal-2024-unsupervised, zhang2025dhcp}, or train a single detector on concatenated features when combining multiple internal states~\cite{nath2025hallushift++, ijcai2025p929}. As a result, they may fail to capture the diverse characteristics of hallucinations. We hypothesize that, since different internal representations capture complementary hallucination signals, combining them can enable more robust hallucination detection.

In this paper, we propose EnsemHalDet, an ensemble-based hallucination detection framework for VLMs that trains independent detectors on attention outputs and hidden states and combines them through detector-level ensemble grounded in ensemble learning theory~\cite{10.1007/3-540-45014-9_1, 10.5555/975251}. Specifically, EnsemHalDet extracts and token-averages the attention outputs and hidden states from the VLMs to serve as features. Then we train individual detectors for each internal state and aggregate them using stacking ensemble.

To evaluate EnsemHalDet, we formulate hallucination detection as a binary classification task using multiple VQA datasets and VLMs. Following the evaluation protocol of CRAG-MM~\cite{crag-mm-2025}, a VQA benchmark, we use an LLM-as-a-judge to compare each generated answer with a human-annotated ground truth and assign a binary label indicating whether the response is hallucinated. We then predict this label using internal representations extracted during answer generation, and evaluate performance using the area under the ROC curve (AUC).

The results show that our framework consistently outperforms existing baselines. Notably, ablation studies confirm that detector-level ensembling provides superior AUC compared to relying on a single representation or simple feature concatenation baselines. In addition, our analysis reveals that different internal representations capture complementary hallucination signals, and effectively combining them improves detection accuracy. EnsemHalDet is also computationally efficient and model-agnostic for open-source models, as it relies on lightweight classifiers while achieving strong performance.

\section{Related Work}

\subsection{Hallucinations in Language and Vision-Language Models}

Hallucination in language models refers to the phenomenon where the model generates fluent and plausible content that is not based on the input or facts~\cite{koehn-knowles-2017-six, maynez-etal-2020-faithfulness, ji2023survey}. 
Hallucinations in VLMs are characterized not only by linguistic errors inherited from language models but also by generations that are inconsistent with the visual input~\cite{10.1145/3703155}.
These categories include object hallucinations referring to nonexistent objects, attribute hallucinations involving incorrect object properties, multi-modal conflicting hallucinations that contradict visual inputs, and counter-common-sense hallucinations that violate visual evidence or common sense~\cite{liu2024phd}.

From the perspectives of reliability and safety, hallucination detection and mitigation are critical research problems. Existing literature explores mitigation via post-hoc training~\cite{11209377, yangzhihe2025mitigating} or latent state interventions~\cite{yang2025mitigating, zhou2025mitigating}. On the other hand, ~\citet{kalai2025language} suggest that hallucinations are an inherent byproduct of optimizing for plausible response generation, making direct mitigation inherently challenging. Accurate detection, however, enables QA systems to adopt an abstention strategy by providing responses such as "I do not know" when uncertainty is detected, thereby significantly improving overall system reliability. Therefore, this work focuses on hallucination detection rather than mitigation.

\subsection{Hallucination Detection Methods}
Existing hallucination detection methods can be broadly categorized into three groups: output-based methods, external-knowledge-based methods, and internal-representation-based methods.

\begin{figure*}[t]
  \centering
  \includegraphics[width=\linewidth]{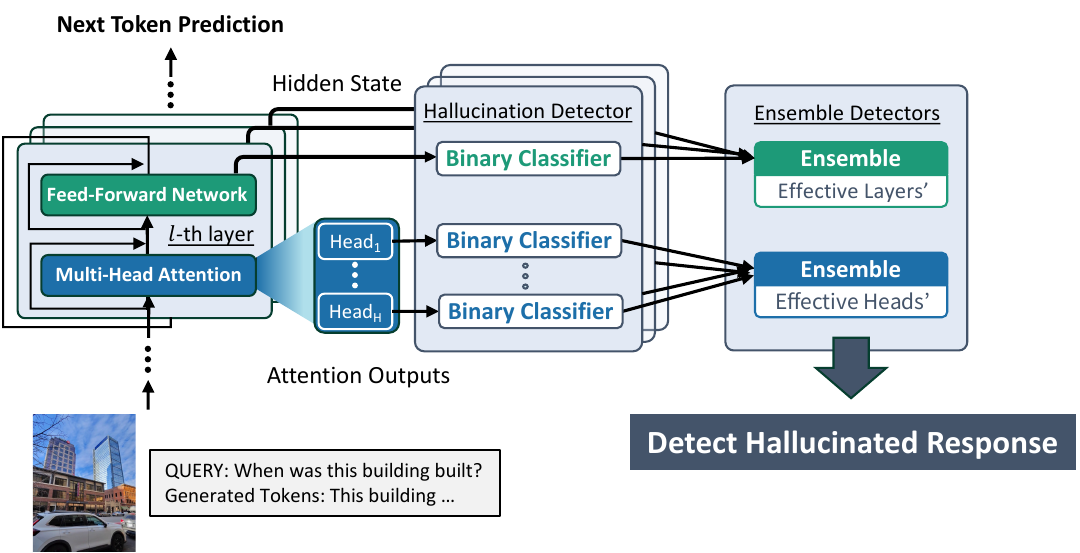}
  \caption{
        Overview of EnsemHalDet: This method extracts attention heads and hidden states across multiple layers. For each representation, binary classifiers are trained via logistic regression to detect hallucinations. By integrating detectors from the most effective layers and heads through ensemble learning, we detect hallucinated answers.
    }
  \label{fig:figure2}
\end{figure*}

Output-based methods detect hallucinations by analyzing generated texts or prediction statistics. These methods analyze model outputs, such as consistency across multiple samples (e.g., SelfCheckGPT~\cite{manakul-etal-2023-selfcheckgpt}) or token-level uncertainty / log-probabilities~\cite{guerreiro-etal-2023-looking, zhang-etal-2023-enhancing-uncertainty}. While model-agnostic, they often fail to detect "confident hallucinations."

External-knowledge-based methods detect hallucinations by verifying generated outputs using external knowledge resources like Wikipedia or web searches to verify facts~\cite{li-etal-2023-halueval, min-etal-2023-factscore}. Although these methods can achieve high detection accuracy, they incur high computational costs and depend heavily on the quality of external data.

Internal-representation-based approaches are based on the premise that a model's internal states encode the truthfulness of its outputs~\cite{li2023inferencetime, marks2023geometry}. Early methods focused on training classifiers using single features (SAPLMA~\cite{azaria-mitchell-2023-internal}, MIND~\cite{su-etal-2024-unsupervised}). Recent studies have expanded this to multiple features, such as cross-attention patterns (DHCP~\cite{zhang2025dhcp}) or the concatenation of hidden states and FFN activations into a single feature vector (MHAD~\cite{ijcai2025p929}). HALLUSHIFT++~\cite{nath2025hallushift++} projects internal states into semantically interpretable statistics, such as layer-wise consistency, attention concentration, confidence, rather than directly concatenating raw internal representations. Internal-representation-based methods tend to achieve higher accuracy than other approaches. However, most existing methods rely on a single internal representation or a single detector, which may limit their ability to capture the diverse characteristics of hallucinations. 

In this work, we exploit multiple internal representations in VLMs, independently train hallucination detectors for each representation and ensemble these detectors to mitigate biases arising from reliance on a single representation or detector.

\section{EnsemHalDet}
 An overview of the proposed method is illustrated in Figure~\ref{fig:figure2}. EnsemHalDet leverages attention head outputs (AH) and hidden states (HS), trains hallucination detectors independently for each attention head and the hidden states of each layer, and then ensembles these detectors. 
 
 \begin{figure*}[t]
  \centering
  \includegraphics[width=\linewidth]{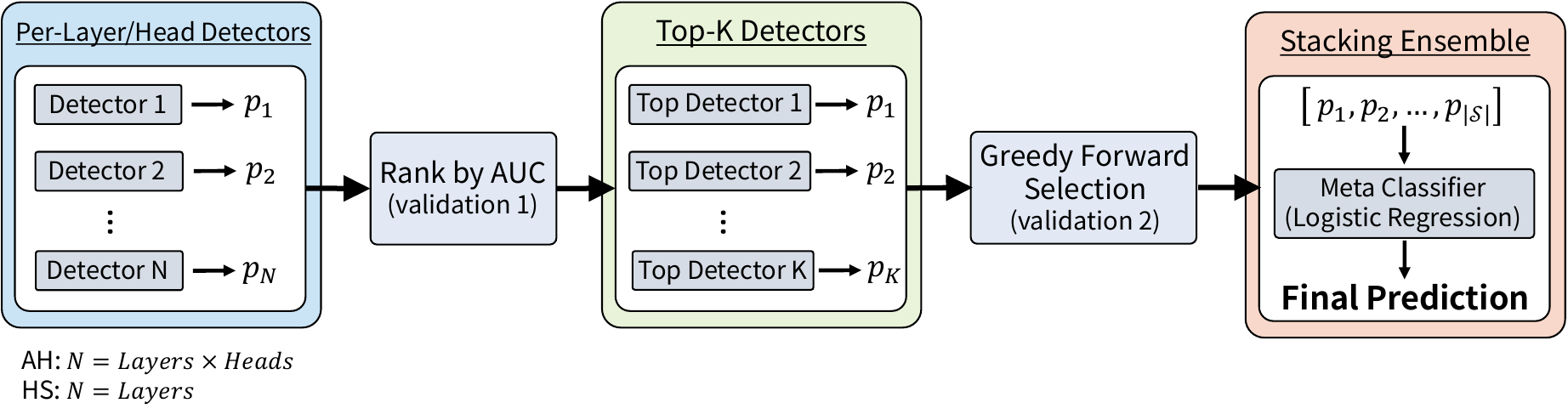}
  \caption{
  Overview of the detector-level ensemble process. For attention-head-based features (AH), we train $L \times H$ detectors, one for each layer-head pair. For hidden-state-based features (HS), we train $L$ detectors, one for each layer. The detectors are first ranked on validation$_1$, filtered to top-K candidates, further refined by greedy forward selection on validation$_2$, and finally combined by a stacking-based meta-classifier (logistic regression).}
  \label{fig:figure3}
\end{figure*}

\subsection{Extract Internal State}
We extract two types of internal representations from the VLM, namely attention outputs and hidden states. Attention outputs are obtained from the multi-head attention module before the output projection and hidden states are taken after the final layer normalization of each transformer block.  Attention representations are extracted at the head level, whereas hidden states are extracted at the layer level.

During generation, at each decoding step $t$, we extract the vector corresponding to the last token from each internal representation, denoted as $\mathbf{e}_t$, which is used for the next token's prediction. We then compute the average over all generated tokens ($T$ tokens),
\begin{equation}
    \bar{\mathbf{e}} = \frac{1}{T}\sum^{T}_{t=1} \mathbf{e}_t, 
\end{equation}
and use this averaged vector as the feature representation for each internal state. The hidden states are compressed to the same dimension as attention outputs using principal component analysis (PCA) because applying PCA consistently improves performance (Appendix Table~\ref{tab:ablation_hs_pca}).

\subsection{Train Detectors}
We train hallucination detectors separately for the attention outputs of each layer and head, as well as for the hidden states of each layer. Given a model with $L$ layers and $H$ attention heads, this results in $L \times H$ attention-based detectors and $L$ hidden-state-based detectors. Each detector is trained via supervised learning using logistic regression, with the extracted internal representations $\bar{\mathbf{e}}$ as input features and a binary label $y$ indicating whether the response is hallucinated:
\begin{equation}
    y=\sigma \left( \mathbf{w_1}^{\top}\bar{\mathbf{e}}+b_1 \right),
\end{equation}
where $\sigma(\cdot)$ denotes the sigmoid function, $\mathbf{w_1}$ is the weight vector, and $b_1$ is the bias term. We use logistic regression due to its simplicity, interpretability, and computational efficiency. All detectors are trained using logistic regression with L2 regularization, a regularization strength of 1.0 ($C$=1.0), and a maximum of 300 optimization iterations. We use the same hyperparameters for all detectors. A detector output probability of 0.5 or higher is regarded as indicating a hallucination.

\subsection{Ensemble Detectors}

Figure~\ref{fig:figure3} illustrates the ensemble pipeline for integrating detectors. We randomly split the validation dataset into two subsets, denoted as validation$_1$ and validation$_2$ and use them to determine the optimal ensemble combination.

First, we evaluate all trained detectors on validation$_1$ and rank them according to AUC. 
Based on this ranking, we select the top-k detectors for each feature type, specifically 30 attention-head-based detectors (AH) and 10 hidden-state-based detectors (HS).

Next, we perform greedy forward selection on validation$_2$.
Starting from an empty set, we iteratively add the detector that yields the largest improvement in AUC. We stop when no further improvement is observed or when the number of detectors reaches 10.

For aggregating the detectors, we adopt stacking ensemble based on preliminary experiments showing superior performance compared to other aggregation methods (Table~\ref{tab:ablation_main}). 
Let $\mathbf{p} = [p_1, p_2, \dots, p_{|\mathcal{S}|}]$ denote the outputs of the selected detectors. 
We train a logistic regression meta-classifier that produces the final prediction as:
\begin{equation}
    p = \sigma \left( \mathbf{w_2}^\top \mathbf{p} + b_2 \right),
\end{equation}
where $\mathbf{w_2}$ and $b_2$ are learnable parameters.

\section{Experimental Setup}
In this section, we present the experimental setup for evaluating EnsemHalDet. We evaluate whether hallucinations in generated answers can be detected from internal states during answer generation, as well as the associated detection time. Specifically, our experiments are designed to assess the effectiveness of EnsemHalDet compared to prior methods, analyze the contribution of different internal representations, examine the impact of ensemble strategies, effect of dimensionality reduction (PCA), and evaluate computational efficiency. 


\subsection{Models, Datasets, and Evaluation Metrics}

\paragraph{Models.}
We conduct experiments using three VLMs: Llama-3.2-11B-Vision-Instruct (Llama-3.2-11B),\footnote{\url{https://huggingface.co/meta-llama/Llama-3.2-11B-Vision-Instruct}} Qwen2.5-VL-7B-Instruct (Qwen2.5-VL-7B),\footnote{\url{https://huggingface.co/Qwen/Qwen2.5-VL-7B-Instruct}} and Pixtral-12B-2409 (Pixtral-12B).\footnote{\url{https://huggingface.co/mistral-community/pixtral-12b}} The architectural details of each model are summarized in Appendix~\ref{appendix:crag_mm_vlm_arch}. 

\paragraph{Datasets.}
As the evaluation dataset, we use CRAG-MM~\cite{crag-mm-2025} and MMMU-Pro~\cite{yue-etal-2025-mmmu}, both of which are VQA benchmarks providing images along with corresponding questions and answers. In this work, we follow the evaluation protocol of CRAG-MM, where the accuracy of generated answers is assessed through an LLM-as-a-Judge framework (GPT-4o-mini\footnote{\url{https://openai.com/index/gpt-4o-mini-advancing-cost-efficient-intelligence/}}). Specifically, the LLM-based judge compares each generated answer against a human-annotated reference to identify potential hallucinations (Prompt: Figure~\ref{fig:judge_prompt}). Following this protocol, we label a generated answer as hallucinated if it is judged to be incorrect compared to the ground truth. Although MMMU-Pro is inherently a multiple-choice benchmark, we reformat it into a free-form generation task and apply the same evaluation protocol as CRAG-MM.

When generating answers to VQA, we do not use a system prompt. Instead, we provide only the image and the question as the user prompt. The CRAG-MM dataset contains 3,874 samples, while MMMU-Pro consists of 1,730 samples. We adapt a stratified split for data splitting to preserve label distributions and divide the dataset into training, validation, and test sets with an 8:1:1 ratio.

\paragraph{Metrics.}
Hallucination detection is formulated as a binary classification task that predicts whether a generated answer is hallucinated using internal representations of VLMs extracted during answer generation. The performance is measured using the area under the ROC curve (AUC), which measures the ability to distinguish between hallucinated and non-hallucinated responses. Detection performance is evaluated as the average over five runs of feature extraction with different random seeds.

We also measure detection time (in seconds) as the sum of two components: (1) feature extraction time and (2) hallucination detection time. All experiments are conducted on a single NVIDIA RTX 6000 Ada GPU with a batch size of 1.

\subsection{Baselines.}
In this study, we compare our method with three model-agnostic hallucination detection approaches based on internal representations: SAPLMA~\cite{azaria-mitchell-2023-internal}, MIND\cite{su-etal-2024-unsupervised}, and MHAD~\cite{ijcai2025p929}, all of which are applicable to open-source language models.
SAPLMA and MIND are early methods that demonstrate the feasibility of detecting hallucinations from internal representations, while MHAD extends this line of work by leveraging multiple internal components. All these methods, including our proposed EnsemHalDet, are model-agnostic for open-source models and can be applied to both VLMs and LLMs.
We use the official implementations for SAPLMA and MIND. Since the implementation of MHAD is not publicly available, we re-implement it based on the descriptions in the original paper.

\paragraph{SAPLMA.}
SAPLMA uses the hidden state of the last token from a specific transformer layer as input features. A 3-layer MLP is used as the classifier.

\paragraph{MIND.}
MIND employs hidden states obtained by concatenating the mean-pooled representation of the final layer and the representation of the last token. A 4-layer MLP is used as the classifier.

\paragraph{MHAD.}
MHAD utilizes outputs from multiple internal components of the model, including attention block outputs (attn), hidden states (hs), and feed-forward networks (ffn).
The first and last tokens from these components are extracted, and a 2-layer MLP is trained to select informative layers and neurons.
The selected representations are concatenated and passed to a 4-layer MLP classifier.

\begin{table*}[t]
\centering
\small
\renewcommand{\arraystretch}{1.2}
\setlength{\tabcolsep}{2.5pt}
\begin{tabular}{l|ccc|ccc}
\toprule
Methods
& \multicolumn{3}{c|}{CRAG-MM}
& \multicolumn{3}{c}{MMMU-Pro} \\
& Llama-3.2-11B & Qwen2.5-VL-7B & Pixtral-12B
& Llama-3.2-11B & Qwen2.5-VL-7B & Pixtral-12B \\
\midrule
SAPLMA
& 0.817 $\pm$ 0.023
& 0.797 $\pm$ 0.023
& 0.730 $\pm$ 0.045
& 0.850 $\pm$ 0.026
& 0.615 $\pm$ 0.076
& 0.791 $\pm$ 0.015 \\
MIND
& 0.822 $\pm$ 0.011
& 0.830 $\pm$ 0.018
& 0.750 $\pm$ 0.021
& 0.866 $\pm$ 0.039
& 0.686 $\pm$ 0.047
& 0.801 $\pm$ 0.034 \\
\midrule
MHAD$_{\text{attn}}$
& 0.850 $\pm$ 0.019
& 0.867 $\pm$ 0.012
& 0.784 $\pm$ 0.012
& 0.870 $\pm$ 0.033
& 0.751 $\pm$ 0.055
& 0.822 $\pm$ 0.034 \\
MHAD$_{\text{hs}}$
& 0.838 $\pm$ 0.028
& 0.849 $\pm$ 0.011
& 0.763 $\pm$ 0.033
& \textbf{0.889 $\pm$ 0.023}
& 0.707 $\pm$ 0.056
& 0.792 $\pm$ 0.046 \\
MHAD$_{\text{ffn}}$
& 0.830 $\pm$ 0.029
& 0.855 $\pm$ 0.020
& 0.777 $\pm$ 0.038
& 0.874 $\pm$ 0.040
& 0.742 $\pm$ 0.050
& 0.810 $\pm$ 0.042 \\
\midrule
EnsemHalDet$_{\text{AH}}$ (ours)
& 0.843 $\pm$ 0.020
& 0.871 $\pm$ 0.021
& 0.800 $\pm$ 0.029
& 0.857 $\pm$ 0.036
& 0.740 $\pm$ 0.036
& 0.802 $\pm$ 0.030 \\
EnsemHalDet$_{\text{HS}}$ (ours)
& \textbf{0.860 $\pm$ 0.013}
& 0.864 $\pm$ 0.012
& 0.817 $\pm$ 0.013
& 0.860 $\pm$ 0.036
& \textbf{0.762 $\pm$ 0.021}
& \textbf{0.826 $\pm$ 0.027} \\
EnsemHalDet$_{\text{MIX}}$ (ours)
& 0.849 $\pm$ 0.021
& \textbf{0.874 $\pm$ 0.019}
& \textbf{0.818 $\pm$ 0.021}
& 0.854 $\pm$ 0.035
& 0.748 $\pm$ 0.037
& 0.802 $\pm$ 0.030 \\
\bottomrule
\end{tabular}
\caption{Main comparison of hallucination detection methods with AUC. All results are reported as the mean and standard deviation over five runs of feature extraction with different random seeds. Best results are in bold.}
\label{tab:main_results_mean_std}
\end{table*}

\subsection{Ablation Study}

We conduct an ablation study to analyze the contribution of each component in EnsemHalDet. 
Specifically, we evaluate the following aspects:

\paragraph{Feature Type.} We compare attention-head-based features (AH), hidden-state-based features (HS), and their combination (MIX).
\paragraph{Combination Strategy.} We compare different strategies for combining multiple detectors, including single detector (top1), feature concatenation (concat), ensemble averaging (average), weighted averaging (weighted), and stacking (stack).
\paragraph{Dimensionality Reduction.} For hidden state based features, we compare performance with and without dimensionality reduction.

\begin{table*}[t]
\centering
\small
\renewcommand{\arraystretch}{1.18}
\setlength{\tabcolsep}{4pt}
\begin{tabular}{l|ccc|ccc}
\toprule
Variants
& \multicolumn{3}{c|}{CRAG-MM}
& \multicolumn{3}{c}{MMMU-Pro} \\
& Llama-3.2-11B & Qwen2.5-VL-7B & Pixtral-12B
& Llama-3.2-11B & Qwen2.5-VL-7B & Pixtral-12B \\
\midrule
\multicolumn{7}{c}{\textbf{Attention-based (AH)}} \\
AH$_{\text{top1}}$
& 0.787 $\pm$ 0.022
& 0.842 $\pm$ 0.029
& 0.744 $\pm$ 0.046
& 0.812 $\pm$ 0.032
& 0.749 $\pm$ 0.025
& 0.737 $\pm$ 0.032 \\
AH$_{\text{concat}}$
& 0.795 $\pm$ 0.031
& 0.833 $\pm$ 0.017
& 0.734 $\pm$ 0.029
& 0.817 $\pm$ 0.047
& 0.689 $\pm$ 0.044
& 0.738 $\pm$ 0.042 \\
AH$_{\text{average}}$
& 0.843 $\pm$ 0.021
& 0.870 $\pm$ 0.021
& 0.801 $\pm$ 0.029
& 0.855 $\pm$ 0.034
& 0.736 $\pm$ 0.035
& 0.803 $\pm$ 0.028 \\
AH$_{\text{weighted}}$
& 0.842 $\pm$ 0.021
& 0.871 $\pm$ 0.023
& 0.805 $\pm$ 0.026
& 0.855 $\pm$ 0.034
& 0.747 $\pm$ 0.040
& 0.797 $\pm$ 0.029 \\
AH$_{\text{stack}}$
& 0.843 $\pm$ 0.020
& 0.871 $\pm$ 0.021
& 0.800 $\pm$ 0.029
& 0.857 $\pm$ 0.036
& 0.740 $\pm$ 0.036
& 0.802 $\pm$ 0.030 \\
\midrule
\multicolumn{7}{c}{\textbf{Hidden-state-based (HS)}} \\
HS$_{\text{top1}}$
& 0.857 $\pm$ 0.015
& 0.863 $\pm$ 0.014
& \textbf{0.818 $\pm$ 0.024}
& 0.845 $\pm$ 0.027
& 0.737 $\pm$ 0.062
& 0.801 $\pm$ 0.030 \\
HS$_{\text{concat}}$
& 0.858 $\pm$ 0.011
& 0.862 $\pm$ 0.013
& \textbf{0.818 $\pm$ 0.015}
& 0.856 $\pm$ 0.040
& 0.743 $\pm$ 0.030
& 0.817 $\pm$ 0.030 \\
HS$_{\text{average}}$
& 0.858 $\pm$ 0.014
& 0.865 $\pm$ 0.012
& 0.816 $\pm$ 0.014
& \textbf{0.860 $\pm$ 0.036}
& 0.758 $\pm$ 0.027
& 0.825 $\pm$ 0.025 \\
HS$_{\text{weighted}}$
& 0.858 $\pm$ 0.014
& 0.865 $\pm$ 0.012
& 0.816 $\pm$ 0.014
& \textbf{0.860 $\pm$ 0.036}
& 0.758 $\pm$ 0.027
& 0.825 $\pm$ 0.026 \\
HS$_{\text{stack}}$
& \textbf{0.860 $\pm$ 0.013}
& 0.864 $\pm$ 0.012
& 0.817 $\pm$ 0.013
& \textbf{0.860 $\pm$ 0.036}
& \textbf{0.762 $\pm$ 0.021}
& \textbf{0.826 $\pm$ 0.027} \\
\midrule
\multicolumn{7}{c}{\textbf{Mixed (AH + HS)}} \\
MIX$_{\text{concat}}$
& 0.842 $\pm$ 0.023
& 0.864 $\pm$ 0.020
& 0.817 $\pm$ 0.016
& 0.857 $\pm$ 0.043
& 0.727 $\pm$ 0.043
& 0.738 $\pm$ 0.042 \\
MIX$_{\text{average}}$
& 0.848 $\pm$ 0.022
& 0.873 $\pm$ 0.019
& 0.813 $\pm$ 0.024
& 0.854 $\pm$ 0.035
& 0.746 $\pm$ 0.038
& 0.803 $\pm$ 0.028 \\
MIX$_{\text{weighted}}$
& 0.847 $\pm$ 0.023
& 0.873 $\pm$ 0.019
& 0.814 $\pm$ 0.026
& 0.853 $\pm$ 0.035
& 0.749 $\pm$ 0.038
& 0.797 $\pm$ 0.029 \\
MIX$_{\text{stack}}$
& 0.849 $\pm$ 0.021
& \textbf{0.874 $\pm$ 0.019}
& \textbf{0.818 $\pm$ 0.021}
& 0.854 $\pm$ 0.035
& 0.748 $\pm$ 0.037
& 0.802 $\pm$ 0.030 \\
\bottomrule
\end{tabular}
\caption{Ablation study of EnsemHalDet. We compare feature types (AH, HS, MIX) and combination strategies (top1, concatenation, average ensemble, weighted ensemble, and stacking).}
\label{tab:ablation_main}
\end{table*}

\begin{table}[t]
\centering
\small
\begin{tabular}{l|ccc}
\toprule
& \multicolumn{3}{c}{CRAG-MM} \\
& Llama & Qwen & Pixtral \\
\midrule
HS$_{\text{top1}}$ (w/ PCA)
& \textbf{0.857} & \textbf{0.863} & \textbf{0.818} \\
HS$_{\text{top1}}$ (w/o PCA)
& 0.755 & 0.775 & 0.721 \\
\midrule
HS$_{\text{stack}}$ (w/ PCA)
& \textbf{0.860} & \textbf{0.864} & \textbf{0.817} \\
HS$_{\text{stack}}$ (w/o PCA)
& 0.774 & 0.782 & 0.750 \\
\bottomrule
\end{tabular}
\caption{Effect of PCA on CRAG-MM for hidden-state-based detectors. PCA consistently improves both single-detector (top1) and ensemble (stack) performance. Full results are provided in Table~\ref{tab:ablation_hs_pca}.}
\label{tab:pca_main}
\end{table}

\section{Results}

\subsection{Comparison with Baselines}
\label{sec:main_results}

Table~\ref{tab:main_results_mean_std} shows the comparison of hallucination detection performance across different methods. All results are reported as the mean and standard deviation over five runs of feature extraction with different random seeds. Overall, EnsemHalDet consistently achieves competitive or superior performance compared to existing approaches across both datasets and all models.

On CRAG-MM, EnsemHalDet outperforms all baselines across all three models, achieving the best AUC in each case. 
For Qwen2.5-VL-7B and Pixtral-12B, the variant that combines both attention-head (AH) and hidden-state (HS) representations achieves the highest performance. 
In contrast, for Llama-3.2-11B, the HS-only variant achieves the best AUC.

On MMMU-Pro, EnsemHalDet achieves the best performance on Qwen2.5-VL-7B and Pixtral-12B, while it does not outperform MHAD on Llama-3.2-11B. 
Additionally, across all three VLMs, the HS-based ensemble consistently outperforms the MIX variant.

These results indicate that EnsemHalDet generalizes well across different models and datasets. Although EnsemHalDet employs a simple logistic regression classifier, its performance suggests that effectively combining multiple internal representations improves robustness and enables accurate hallucination detection.

\subsection{Ablation Study}
\label{sec:ablation}

Table~\ref{tab:ablation_main} presents the ablation study results of EnsemHalDet. All results are reported as the mean and standard deviation over five runs of feature extraction with different random seeds. 

\paragraph{Effect of Feature Types.}
First, we analyze the impact of different internal representations. 
Across most settings, hidden-state-based (HS) detectors consistently outperform attention-based (AH) detectors, indicating that hidden states provide more informative signals for hallucination detection. 
This trend is particularly evident on CRAG-MM, where HS-based methods achieve higher AUC than AH-based methods for all three models.

Combining both representations (MIX) further improves performance in some cases, especially for Qwen2.5-VL-7B and Pixtral-12B on CRAG-MM.  However, this improvement is not consistent across all settings.  In particular, HS-only variants often outperform MIX on MMMU-Pro, suggesting that combining multiple feature types does not always lead to better performance. 

This implies that the benefit of combining representations depends on the model and dataset, and that HS features alone already capture strong hallucination signals.

\paragraph{Effect of Combination Strategies.}
Next, we examine different combination strategies. 
Ensemble-based methods (average, weighted, and stack) consistently outperform single-detector (top1) and feature-concatenation (concat) approaches across all settings. In particular, ensemble methods lead to substantial performance improvements in AH-based hallucination detection.
This demonstrates the importance of leveraging multiple detectors rather than relying on a single representation or simply merging features.
Among the ensemble methods, stacking achieves the best overall performance in most cases. 
This indicates that learning how to combine detector outputs is more effective than simple aggregation strategies such as averaging or concatenation.

\paragraph{Effect of PCA for Hidden States.}
We further analyze the effect of PCA on hidden-state-based detectors in Table~\ref{tab:pca_main}. Applying PCA consistently improves performance for most combination strategies, particularly for top1 and ensemble-based methods. This suggests that dimensionality reduction helps remove noise and improves the stability of learned detectors. 
In addition, the dimensionality of hidden-state features is substantially larger than the number of training samples, which may lead to overfitting and unstable optimization. By reducing the feature dimensionality, PCA alleviates this issue and enables more effective learning.

\subsection{Inference Time Analysis}

\begin{table*}[t]
\centering
\begin{tabular}{lcccc}
\toprule
Method & Feature Extraction (s) & Detection (s) & Total (s) & Percentage (\%) \\
\midrule
Baseline & 2.511 & -- & 2.511 & 100.0 \\
\midrule
SAPLMA & 3.767 & 0.000060 & 3.767 & 150.0 \\
MIND & 5.591 & 0.001465 & 5.592 & 222.7 \\
MHAD$_{\text{attn}}$ & 3.095 & 0.001835 & 3.097 & 123.3 \\ 
MHAD$_{\text{hs}}$ & 2.952 & 0.002083 & 2.954 & 117.7 \\ 
MHAD$_{\text{ffn}}$ & 3.182 & 0.001325 & 3.183 & 126.8 \\ 
\midrule
EnsemHalDet$_{\text{AH}}$ (ours) & 3.095 & 0.000023 & 3.095 & 123.2 \\
EnsemHalDet$_{\text{HS}}$ (ours) & 4.618 & 0.000054 & 4.618 & 183.9 \\
EnsemHalDet$_{\text{MIX}}$ (ours) & 5.220 & 0.000120 & 5.220 & 207.9 \\
\bottomrule
\end{tabular}
\caption{Inference time comparison for hallucination detection. The percentage indicates the total runtime relative to the baseline generation time (100\%). We report GPU-based inference results for SAPLMA, MIND, and MHAD.}
\label{tab:runtime_main}
\end{table*}

We analyze the inference time of each method by decomposing it into feature extraction and hallucination detection. 

Overall, the feature extraction stage dominates the total runtime across all methods. In contrast, the detection stage is negligible, typically taking less than 0.002 seconds. This indicates that the computational bottleneck lies in extracting internal representations rather than in classifier inference.

Our proposed method, EnsemHalDet, introduces additional overhead due to the use of multiple representations. Specifically, EnsemHalDet$_{\text{AH}}$ achieves a runtime comparable to MHAD$_{\text{attn}}$ (3.095s), while EnsemHalDet$_{\text{HS}}$ and EnsemHalDet$_{\text{MIX}}$ require longer processing time due to the aggregation of multiple features.

Compared to baseline methods, EnsemHalDet incurs higher computational cost, particularly for the HS and MIX variants. This is mainly due to our design choice of extracting and aggregating representations from all generated tokens, whereas prior methods such as SAPLMA and MHAD only utilize limited tokens (e.g., the first and last tokens).

Despite the increased feature extraction cost, the detection stage of EnsemHalDet remains highly efficient. In particular, EnsemHalDet achieves significantly faster detection, being roughly 20--80× faster than MHAD. Although EnsemHalDet does not always outperform MHAD in terms of detection accuracy, it achieves substantially faster detection time.
This is because our method employs a lightweight logistic regression classifier, whereas prior methods rely on MLPs.

These results highlight a trade-off between computational cost and detection performance. While EnsemHalDet introduces additional overhead in feature extraction, it maintains efficient detection and achieves superior performance, making it a practical approach for hallucination detection. We further observe that the AH- and HS-based ensemble variants already achieve sufficiently strong performance. Considering the additional cost of extracting multiple types of representations, using detectors trained on a single type of representation and ensembling them can provide a more efficient and practical alternative.

\section{Conclusion}

In this paper, we proposed EnsemHalDet, an ensemble-based hallucination detection framework for vision-language models that ensembles multiple detectors trained on diverse internal representations. 
Experimental results on CRAG-MM and MMMU-Pro across multiple VLMs demonstrate that EnsemHalDet consistently outperforms existing baselines in terms of detection AUC. Through extensive ablation studies, we showed that hidden-state-based features provide strong signals for hallucination detection, and that ensemble-based approaches, particularly stacking, more effectively detect hallucinations compared to single-feature detectors and feature-concatenation methods. 
These findings highlight the importance of leveraging multiple internal representations to improve robustness and accuracy in hallucination detection.
While our method achieves strong performance, it incurs additional computational cost due to the extraction of multiple internal states. Addressing this efficiency--performance trade-off remains an important direction for future work.
In future work, we plan to explore more efficient feature selection strategies, extend our framework to standard large language models, and investigate why ensemble-based methods are particularly effective for hallucination detection.

\section*{Limitations}
EnsemHalDet is model-agnostic for open-source VLMs and can be flexibly applied across a wide range of architectures. However, it requires access to internal representations, which limits its applicability to closed-source or API-only models.

Second, while our experiments show that detector-level ensembling outperforms single-detectors and feature concatenation, we do not provide a theoretical or empirical analysis explaining why ensembling is more effective than feature concatenation in this setting.

Finally, hallucination labels are determined using an LLM-as-a-Judge framework, and we do not analyze how the effectiveness of detection varies across different types of hallucinations.

\section*{Acknowledgments}
This work was supported by JST K Program Japan Grant Number JPMJKP24C3.


\bibliography{custom}

\newpage
\clearpage

\appendix

\begin{table*}[!htbp]
\centering
\small
\begin{tabular}{lccccc}
\toprule
Model & \#Layers & \#Heads & Hidden Dim & Multimodal Fusion & Hallucination Rate \\
\midrule
Llama-3.2-11B-Vision-Instruct & 40 & 32 & 4096 & Cross-attention & 0.795 \\
Qwen2.5-VL-7B-Instruct & 28 & 28 & 3584 & Projector + Self-attention & 0.837 \\
Pixtral-12B-2409 & 40 & 32 & 4096 & Projector + Self-attention & 0.823 \\
\bottomrule
\end{tabular}
\caption{Architectural details of VLMs, including hidden state dimensionality.}
\label{tab:vlm_arch}
\end{table*}

\section{VLMs architectures}
\label{appendix:crag_mm_vlm_arch}

Table~\ref{tab:vlm_arch} shows the architectures of each VLM that we used in the experiments.
Llama-3.2-11B-Vision-Instruct integrates multimodal information by computing cross-attention between image and text representations at specific layers. 
In contrast, Qwen2.5-VL-7B-Instruct and Pixtral-12B-2409 project visual features into the same embedding space as text, treating image tokens equivalently to text tokens and integrating multimodal information via self-attention.
The evaluated models exhibit relatively high hallucination rates, ranging from 79.5\% to 83.7\%. This is primarily attributed to the challenging nature of zero-shot VQA on the CRAG-MM benchmarks, particularly as we do not employ any system prompts to guide the model's behavior.

\section{LLM-as-a-Judge Prompt}
\label{appendix:judge_prompt}

 Following the evaluation protocol of CRAG-MM~\cite{crag-mm-2025}, we use GPT-4o-mini as the evaluator to judge whether the generated answer is hallucinated. Using the same evaluator ensures consistency with prior work and allows for fair comparison across methods. The exact prompt used in our experiments is shown below.
 
\begin{figure}[h]
\centering
\includegraphics[width=0.9\linewidth]{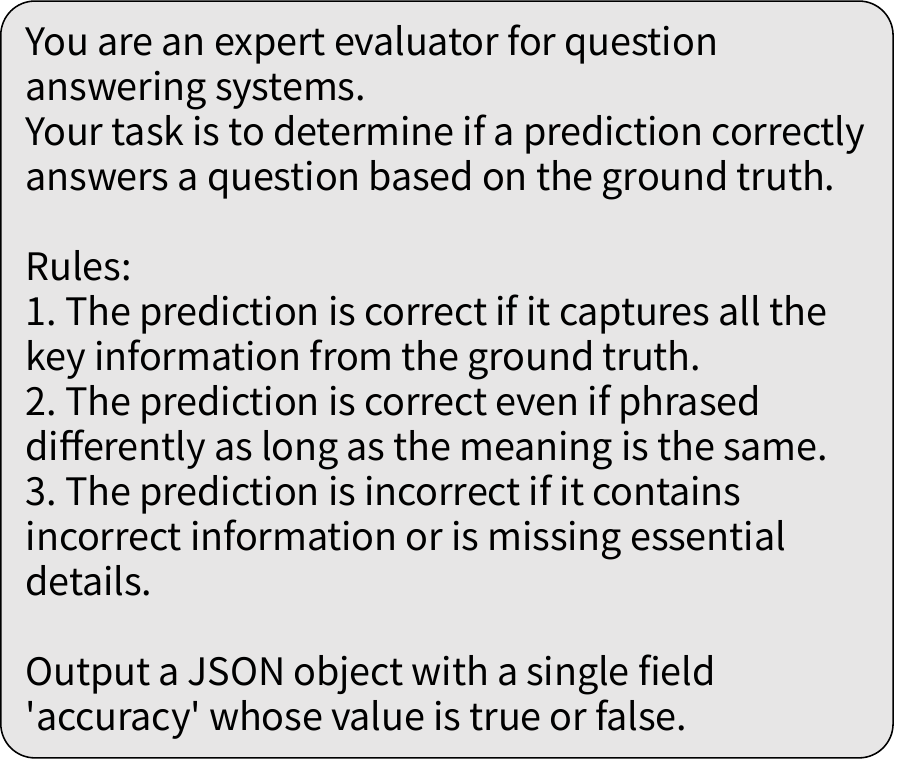}
\caption{Prompt used for hallucination evaluation. We follow the CRAG-MM evaluation protocol.}
\label{fig:judge_prompt}
\end{figure}

\section{Effect of PCA}
\label{appendix:pca}

\begin{table*}[t]
\centering
\small
\renewcommand{\arraystretch}{1.15}
\setlength{\tabcolsep}{4pt}
\begin{tabular}{l|ccc|ccc}
\toprule
Variants
& \multicolumn{3}{c|}{CRAG-MM}
& \multicolumn{3}{c}{MMMU-Pro} \\
& Llama-3.2-11B & Qwen2.5-VL-7B & Pixtral-12B
& Llama-3.2-11B & Qwen2.5-VL-7B & Pixtral-12B \\
\midrule
HS$_{\text{top1}}$ (w/ PCA)
& \textbf{0.857 $\pm$ 0.015}
& \textbf{0.863 $\pm$ 0.014}
& \textbf{0.818 $\pm$ 0.024}
& \textbf{0.845 $\pm$ 0.027}
& \textbf{0.737 $\pm$ 0.062}
& \textbf{0.801 $\pm$ 0.030} \\
HS$_{\text{top1}}$ (w/o PCA)
& 0.755 $\pm$ 0.023
& 0.775 $\pm$ 0.019
& 0.721 $\pm$ 0.023
& 0.807 $\pm$ 0.076
& 0.657 $\pm$ 0.063
& 0.761 $\pm$ 0.065 \\
\midrule
HS$_{\text{average}}$ (w/ PCA)
& \textbf{0.858 $\pm$ 0.014}
& \textbf{0.865 $\pm$ 0.012}
& \textbf{0.816 $\pm$ 0.014}
& \textbf{0.860 $\pm$ 0.036}
& \textbf{0.758 $\pm$ 0.027}
& \textbf{0.825 $\pm$ 0.025} \\
HS$_{\text{average}}$ (w/o PCA)
& 0.774 $\pm$ 0.022
& 0.782 $\pm$ 0.020
& 0.750 $\pm$ 0.020
& 0.816 $\pm$ 0.073
& 0.682 $\pm$ 0.067
& 0.784 $\pm$ 0.051 \\
\midrule
HS$_{\text{stack}}$ (w/ PCA)
& \textbf{0.860 $\pm$ 0.013}
& \textbf{0.864 $\pm$ 0.012}
& \textbf{0.817 $\pm$ 0.013}
& \textbf{0.860 $\pm$ 0.036}
& \textbf{0.762 $\pm$ 0.021}
& \textbf{0.826 $\pm$ 0.027} \\
HS$_{\text{stack}}$ (w/o PCA)
& 0.774 $\pm$ 0.022
& 0.782 $\pm$ 0.020
& 0.750 $\pm$ 0.020
& 0.816 $\pm$ 0.073
& 0.682 $\pm$ 0.067
& 0.784 $\pm$ 0.051 \\
\bottomrule
\end{tabular}
\caption{Ablation study on the effect of PCA for hidden-state-based detectors. Applying PCA consistently improves performance for top1, average ensemble, weighted ensemble, and stacking, while the effect on concatenation is mixed.}
\label{tab:ablation_hs_pca}
\end{table*}

We further analyze the effect of PCA on hidden-state-based detectors in Table~\ref{tab:ablation_hs_pca}. 
Applying PCA consistently improves performance for most combination strategies, particularly for top1 and ensemble-based methods. 
This suggests that dimensionality reduction helps remove noise and improves the stability of learned detectors. 
In contrast, the effect on feature concatenation is less consistent, likely due to differences in how high-dimensional features are utilized.

\section{Ablation of Inference Time}
\label{appendix:inference_time}

\begin{table*}[t]
\centering
\begin{tabular}{lcccc}
\toprule
Method & Feature Extraction (s) & Detection (s) & Total (s) & Percentage (\%) \\
\midrule
Baseline & 2.511 & -- & 2.511 & 100.0 \\
\midrule
SAPLMA (CPU) & 3.767 & 0.000219 & 3.767 & 150.0 \\
SAPLMA (GPU) & 3.767 & 0.000060 & 3.767 & 150.0 \\

MIND (CPU) & 5.591 & 0.001465 & 5.592 & 222.7 \\
MIND (GPU) & 5.591 & 0.001430 & 5.592 & 222.7 \\

MHAD$_{\text{attn}}$ (CPU) & 3.095 & 0.001835 & 3.097 & 123.3 \\
MHAD$_{\text{attn}}$ (GPU) & 3.095 & 0.001309 & 3.096 & 123.3 \\ 

MHAD$_{\text{hs}}$ (CPU) & 2.952 & 0.002083 & 2.954 & 117.7 \\
MHAD$_{\text{hs}}$ (GPU) & 2.952 & 0.001412 & 2.954 & 117.7 \\ 

MHAD$_{\text{ffn}}$ (CPU) & 3.182 & 0.001325 & 3.183 & 126.8 \\
MHAD$_{\text{ffn}}$ (GPU) & 3.182 & 0.001465 & 3.184 & 126.8 \\ 

\midrule
EnsemHalDet$_{\text{AH}}$ (ours) & 3.095 & 0.000023 & 3.095 & 123.2 \\
EnsemHalDet$_{\text{HS}}$ (ours) & 4.618 & 0.000054 & 4.618 & 183.9 \\
EnsemHalDet$_{\text{MIX}}$ (ours) & 5.220 & 0.000120 & 5.220 & 207.9 \\
\bottomrule
\end{tabular}
\caption{Detailed runtime breakdown. Feature extraction corresponds to internal representation extraction, 
while hallucination detection denotes classifier inference time.}
\label{tab:runtime_appendix}
\end{table*}

Table~\ref{tab:runtime_appendix} presents a detailed breakdown of inference time across different methods.

We compared CPU-based and GPU-based inference for prior methods. 
Across SAPLMA, MIND, and MHAD, the difference between CPU and GPU execution is minimal, as the detection stage accounts for only a small fraction of the total runtime. 
This indicates that accelerating the classifier alone provides limited benefit, since the overall cost is dominated by feature extraction.

\end{document}